%% file: arxiv.tex
\documentclass{article}

\usepackage{PRIMEarxiv}

\usepackage{graphicx}
\usepackage{hyperref}
\usepackage[dvipsnames]{xcolor}
\usepackage{todonotes}
\usepackage{booktabs}
\usepackage{microtype}
\usepackage{graphicx}
\usepackage{amsfonts}
\usepackage{adjustbox}
\usepackage{array}
\usepackage{multirow}

\usepackage{pifont}

\newcommand\mc{\mathcal}

\usepackage{listings}
\usepackage{xcolor}       
\usepackage{multirow}
\usepackage{caption} 

\usepackage{pifont}

\newcolumntype{P}[1]{>{\centering\arraybackslash}p{#1}}

\colorlet{lightyellow}{yellow!40}

\makeatletter
{\small 
\xdef\f@size@small{\f@size}
\xdef\f@baselineskip@small{\f@baselineskip}
\normalsize 
\xdef\f@size@normalsize{\f@size}
\xdef\f@baselineskip@normalsize{\f@baselineskip}
}
\newcommand{\smalltonormalsize}{%
  \fontsize
    {\fpeval{(\f@size@small+\f@size@normalsize)/2}}
    {\fpeval{(\f@baselineskip@small+\f@baselineskip@normalsize)/2}}%
  \selectfont
}
\makeatother

\title{Marking: Visual Grading with \\ Highlighting Errors and Annotating Missing Bits}

\author{
  Shashank Sonkar\thanks{Equal contribution.}\\
  Rice University \\
  Houston, TX \\
  \texttt{ss164@rice.edu} \\
  \And
  Naiming Liu\footnotemark[1]\\
  Rice University \\
  Houston, TX \\
  \texttt{nl35@rice.edu} \\
  \AND
  Debshila B. Mallick\\
  Rice University \\
  Houston, TX \\
  \texttt{db19@rice.edu} \\
  \And
  Richard G. Baraniuk \\
  Rice University \\
  Houston, TX \\
  \texttt{richb@rice.edu} \\
}

\begin{document}
\maketitle

\begin{abstract}
\input{sections/abstract}

\end{abstract}

\section{Introduction}
\input{sections/intro}

\section{Related Work}
\input{sections/related_work}

\section{Problem Formulation}
\input{sections/methodology}

\section{Dataset}

\input{sections/dataset}

\section{Experiments and Baselines}
\input{sections/benchmark}

\section{Conclusion}
\input{sections/conclusion}

\section*{Acknowledgments}
This work was supported by NSF grants 1842378, ONR grant N0014-20-1-2534, AFOSR grant FA9550-22-1-0060, and a Vannevar Bush Faculty Fellowship, ONR grant N00014-18-1-2047.

\bibliographystyle{unsrt}  
\bibliography{mybibliography}

\end{document}

%% file: sections/abstract.tex
In this paper, we introduce ``Marking", a novel grading task that enhances automated grading systems by performing an in-depth analysis of student responses and providing students with visual highlights. 
Unlike traditional systems that provide binary scores, ``marking" identifies and categorizes segments of the student response as correct, incorrect, or irrelevant and detects omissions from gold answers. 
We introduce a new dataset meticulously curated by Subject Matter Experts specifically for this task. 
We frame ``Marking" as an extension of the Natural Language Inference (NLI) task, which is extensively explored in the field of Natural Language Processing. 
The gold answer and the student response play the roles of premise and hypothesis in NLI, respectively. 
We subsequently train language models to identify entailment, contradiction, and neutrality from student response, akin to NLI, and with the added dimension of identifying omissions from gold answers.
Our experimental setup involves the use of transformer models, specifically BERT and RoBERTa, and an intelligent training step using the e-SNLI dataset. 
We present extensive baseline results highlighting the complexity of the ``Marking" task, which sets a clear trajectory for the upcoming study. 
Our work not only opens up new avenues for research in AI-powered educational assessment tools, but also provides a valuable benchmark for the AI in education community to engage with and improve upon in the future.
The code and dataset can be found at \url{https://github.com/luffycodes/marking}.

%% file: sections/intro.tex
Recent advancements in Artificial Intelligence (AI) have revolutionized the approach to automated grading in educational settings, offering scalable solutions for evaluating student work\cite{valenti2003overview}. Despite these advances, most systems prioritize binary, right-or-wrong assessments, lacking the subtlety and depth of feedback that educators typically provide. Recognizing this limitation, our work introduces a new dimension to automated grading, moving towards a more detailed form of assessment.

We introduce ``Marking", an innovative task designed to push the capabilities of automated grading systems beyond their current scope. `Marking' leverages AI to perform an in-depth analysis of student responses by evaluating them against a gold standard. The task involves training a language model to meticulously highlight text segments in the student's response, categorizing them as correct, incorrect, or irrelevant. Concurrently, it includes identifying and flagging content in the gold standard that the student has omitted. This two-fold assessment approach ensures a rigorous examination of what students express in their answers and highlights what they fail to include, offering a detailed and balanced view of their knowledge and understanding. By doing so, ``Marking" provides a more nuanced insight into student learning, comparable to the detailed feedback typically provided by experienced educators.

\input{figures/marking_demo}

In our ``Marking" framework, the evaluation of student responses draws on principles similar to Natural Language Inference (NLI) \cite{dagan2006pascalrte}, a core task in Natural Language Processing (NLP). Traditionally, in NLI, a hypothesis (the student response) is tested against a premise (the gold answer) to determine if it is true (entailment), false (contradiction), or indeterminate (neutral). In the context of `Marking,' we segment the student response into three parts: ones that accurately reflect the gold answer is tagged as ``entailment", ones that misrepresent the gold answer as ``contradiction," and those unrelated to the question as ``neutral."

To evaluate the effectiveness of our ``Marking" framework, we conduct a series of experiments using transformer models, specifically BERT \cite{bert} and RoBERTa \cite{liu2019roberta}, which have shown remarkable performance in various NLP tasks.
We treat the task of ``Marking" as a classification problem, where the objective is to label the words in the student response as correct, incorrect, or irrelevant based on the gold standard answer. 
In our experimental setup, we fine-tune the aforementioned masked language models by adding an inference label classifier on top, thereby adapting these models to our specific task of marking.
Our experiments investigate the impact of two preprocessing steps, Dual Instance Pairing (DIP) and stopword removal, and establish baselines for different label settings. 
Our results indicate that both DIP and stopword removal contribute positively to the model performance, which provides valuable insights into the potential of our ``Marking" framework for enhancing automated grading systems.


In pursuit of realizing such an AI-driven marking system, this paper makes the following contributions to the field of AI in education:
\begin{enumerate}
    \item We conceptualize and define the task of ``Marking," delineating its scope and significance as a more granular and informative assessment method than currently prevalent automated grading approaches.

    \item We introduce a novel dataset, BioMarking, meticulously curated by biology Subject Matter Experts, specifically designed to train and evaluate AI models on the ``Marking" task. This dataset also serves as a benchmark for the community in the upcoming study.
    
    \item Through extensive experiments on our dataset, we provide the baseline performance of our ``Marking" task with various settings. This baseline results highlight the complexity and challenge of ``Marking", thus setting a clear trajectory for future research.
\end{enumerate}

The rest of the paper begins with a review of related work that contextualizes our approach within the existing landscape of AI-powered grading tools. Subsequent sections detail the problem statement, dataset creation, experimental results, and a discussion that collectively underscores the potential of ``Marking" to transform AI-based educational assessment.

%% file: figures/marking_demo.tex
\begin{figure*}[t!]
\centering
  \includegraphics[width=\textwidth]{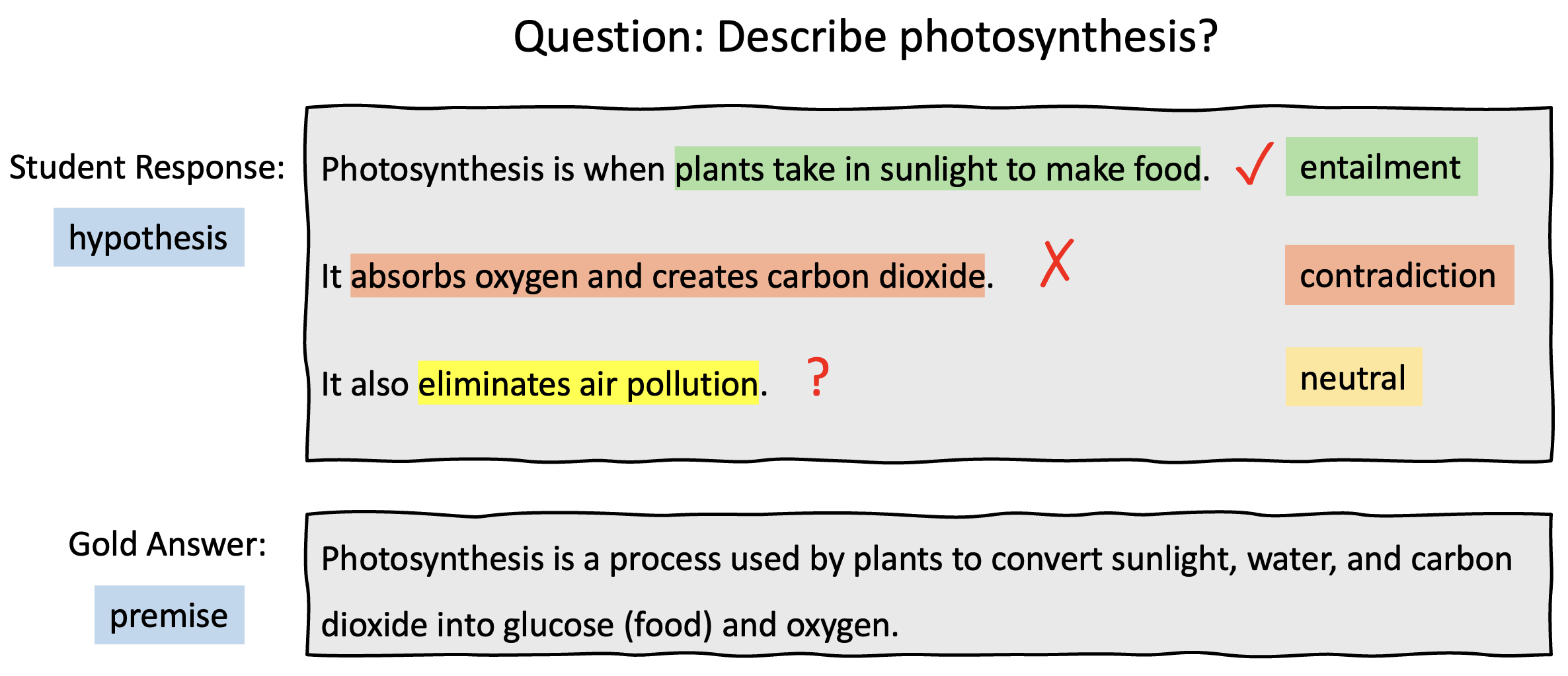}
  \caption{An illustration of the ``Marking" task, which is formulated as NLI where the ``gold answer" represents the premise and the ``student response" is the hypothesis. The correct parts of the student response is classified as \textit{entailment}, the incorrect parts as \textit{contradiction}, and irrelevant part as \textit{neutral}. }
  \label{fig:example}
  \vspace{-10pt}
\end{figure*}

%% file: sections/related_work.tex
\subsection{Automated Grading and Feedback}
LLMs have made a significant impact in the field of AI-enhanced education \cite{sonkar2023class,sonkar2024code,sonkar2024pedagogical} in recent years.
One notable area where LLMs have been applied is automated grading.
Automated grading systems have gained popularity in recent years, especially with the advent of natural language processing and transformer-based models~\cite{attention} such as BERT~\cite{bert} and GPT models~\cite{gpt-3,gpt-2}. 
Furthermore, the integration of automated feedback generation with grading in education has also accelerated~\cite{cavalcanti2019analysis,ramesh2022automated}. Such automated feedback generation approaches are usually applied to structured questions that require well-defined answers, such as multiple-choice questions~\cite{narayanan2017question}, fill-in-the-blank~\cite{sahu2019feature}, or programming assignments~\cite{parihar2017automatic}. However, this automated feedback has some drawbacks as it increases the complexity of measuring the feedback quality compared to manual graders~\cite{kurdi2020systematic}. Our novel ``Marking" task can resolve the challenge of feedback generation as it provides immediate and easy-to-follow feedback to students directly based on their responses, without requiring extra steps to evaluate the quality of the feedback.

\subsection{Natural Language Inference}

Natural language inference (NLI) is the task of determining whether a ``hypothesis" is true (entailment), false (contradiction), or undetermined (neutral) given a ``premise." 
Early research in the field of NLI focused on handcrafted features and classic machine learning techniques~\cite{wang2015learning,chen2016enhanced}. Models using these classical methods tend to require extensive feature engineering and domain knowledge. 
The integration of transformer models with NLI~\cite{kumar2020nile,bert} has effectively gained considerable attention in recent years, which offers a significant enhancement over former methods. 
NLI has broad applications, such as information retrieval, semantic parsing, and commonsense reasoning~\cite{bos2005recognising,maccartney2009extended}.
In this paper, we frame the ``Marking" approach as NLI tasks, which perceives the incorrect portion of the student response as ``contradiction" against the gold standard answer, while the correct portion as ``entailment". This approach characterizes the possibilities for a more dynamic grading system, where graders can denote parts of students' responses as either correct or incorrect, generating detailed and layered feedback.

%% file: sections/methodology.tex
\subsection{Problem Statement}

Given a dataset \( \mathcal{D} = \{ (R_i, A_i, G_i) \}_{i=1}^{N} \), where \( R_i \) represents a student's response, \( A_i \) represents the annotations for the correct, incorrect, and irrelevant portions of \( R_i \), and \( G_i \) is the gold answer for the \( i^{th} \) question, the task of ``Marking" is to learn a function \( f: (R, G) \rightarrow A \) that predicts the annotations \( A \) for a pair of student response \( R \) and gold  answer \( G \).

The annotation \( A_i \) is structured as follows:

\begin{itemize}
    \item \( A_i^{entail} \subseteq R_i \): The subset of the student's response that correctly aligns with the information in the gold answer, representing ``entailment".
    \item \( A_i^{contradict} \subseteq R_i \): The subset of the student's response that incorrectly contradicts the information in the gold answer, representing ``contradiction".
    \item \( A_i^{neutral} \subseteq R_i \): The subset of the student's response that is irrelevant or tangential to the gold answer, representing ``neutral".
\end{itemize}

Additionally, the task requires the identification of omissions from gold answer:
\begin{itemize}
    \item \( A_i^{omission} \subseteq G_i \): The subset of the gold standard that contains key concepts or information not present in the student's response, representing ``omission".
\end{itemize}

The objective of ``Marking" is to optimize the function \( f \) based on a suitable performance metric, such as F1 score, across the annotations. Formally, the F1 score for each category is defined as:

\[ F1_{category} = 2 \times \frac{Precision_{category} \times Recall_{category}}{Precision_{category} + Recall_{category}} \]

where \( category \) can be `entailment', `contradiction', `neutral', or `omission', and \( Precision \) and \( Recall \) are calculated based on the true positives (TP), false positives (FP), and false negatives (FN) for each category:

\[ Precision_{category} = \frac{TP_{category}}{TP_{category} + FP_{category}} \]

\[ Recall_{category} = \frac{TP_{category}}{TP_{category} + FN_{category}} \]

The overall performance of the ``Marking" system could be evaluated using the precision, recall and the weighted average of F1 scores across the different categories, accounting for their relative importance or frequency in the dataset.

\subsection{Neutral condition in Omissions}
In the standard NLI tasks, the ``neutral" category is used to classify statements (hypotheses) that neither clearly support (entail) nor contradict the reference statement (premise); they are essentially unrelated or cannot be verified based on the given information. In the `Marking" task, the ``neutral" category is repurposed. It is no longer just a label for content that is neither right nor wrong in the student's response. Instead, it is now actively used to identify key concepts or information that are present in the gold answer but are missing from the student's response. In other words, ``neutral" in this context is being used to flag omissions — important elements that the student did not include in their answer but should have, according to the gold answer.

By redefining the ``neutral" category in this way, the AI system can provide feedback that is not only about what the student wrote but also about what the student failed to write. This proactive use of the ``neutral" category effectively transforms it into a diagnostic tool that can provide comprehensive feedback to students, pointing out gaps in their knowledge or understanding that need to be addressed.

To summarize, the ``neutral" category is being proactively used to enhance the depth of analysis in grading by identifying and highlighting missing elements in student responses, thereby offering a more complete and informative assessment.

%% file: sections/dataset.tex
We developed the BioMarking dataset to fill in the blanks of the current ``Marking" tasks. Our dataset contains 11 college-level biology questions along with 318 student responses. In the remainder of this section, we report the dataset construction process and its key statistics.

\subsection{Dataset Construction}

\noindent\textbf{~Question Selection}~
We select 11 biology questions from OpenStax \footnote{OpenStax: https://openstax.org/} test bank with college-level multiple choice questions. The tutoring system adopted by OpenStax is that the students first answer the biology multiple-choice questions as if they were free-response questions without seeing the choices. Then the system displays the choices and the students pick the answer from the multiple-choice options. The two-step question-answering strategy is designed to help the students learn the materials better. In order to ensure the quality of our dataset, we only extract the responses of the students who select incorrect multiple choices. Additionally, we carefully curate the gold answer of each question based on the correct multiple-choice option, explanation for each choice, and solution to ascertain its accuracy and comprehensiveness. These answers are then thoroughly reviewed and approved by experts in the field of biology to ensure their validity.

\input{tables/dataset}

\vspace{2mm}
\noindent\textbf{~Annotation Process}~
The annotation task was entrusted to a dedicated team of six Subject Matter Experts (SMEs). All of the SMEs have graduate-level education in biology. Leveraging their respective depths of understanding and expertise, they engaged in the meticulous process of marking all the student answers based on the provided gold answers. Each question is graded by two SMEs simultaneously. We instruct the SMEs to try to avoid labeling the whole sentence, but only the phrases that are incorrect. We provide three types of labels for the SMEs to identify in the student responses: 
\begin{itemize}

  \item Use \textless \textgreater \ to identify the \textbf{correct parts} and color them in \textcolor{ForestGreen}{green}.

  \item Use [] to identify the \textbf{incorrect parts} and color them in \textcolor{red}{red}.

  \item Use \{\} to identify \textbf{irrelevant parts} and color them in \textcolor{Dandelion}{yellow}. 

\end{itemize}

where the correct/incorrect part refers to the section of student answers that agree/contradict with what \textbf{actually present} in the provided gold answers. The definition of irrelevant is the information in the student response unrelated to the question or the solution, \textbf{regardless of the correctness}. 

We also provide a type of label to mark the solutions: 
\begin{itemize}

  \item Use \{\} to highlight parts in the gold answer that are \textbf{missing} in the student response and color them in \textcolor{blue}{blue}.

\end{itemize}

In addition to marking the sections of the student responses, we also ask the SMEs to provide a grade from \{0, 0.5, 1\} to evaluate whether the responses correctly address the solution, where 0 means the student response is incorrect, 0.5 means partially correct and 1 means correct response. We also recommend the SMEs provide some feedback on why the markings are given only based on incorrect parts, such as a misconception of a fact that leads to students’ incorrect reasoning or a counterfactual statement, etc. Examples of the annotated dataset can be found in Table~\ref{dataset}.

\vspace{2mm}
\noindent\textbf{~Annotator training}~
We provide the SME annotators with detailed instructions on the grading task. To demonstrate their understanding and application of these instructions, they were first given a practice set of three questions and six student responses. Upon receiving their completed practice sets, we reviewed the marked responses for consistency and accuracy. We identify the discrepancies based on the gold answer created by the expert supervisors and provide them with feedback. The process was essential to ensure minimal disagreement between annotators during the actual grading process.

\subsection{Dataset Statistics}

The BioMarking dataset comprises 11 college-level biology questions with 318 distinct student responses. Two SME annotators grade each question, and we provide both annotated student responses to ensure the completeness of our dataset.. Each question received an average of 29 student responses, with a standard deviation of 7.6. It is worth noting that the questions are initially presented as multiple-choice, with an average accuracy rate of 82.8\%. However, individual questions varied in difficulty, as demonstrated by a minimum accuracy rate of 55.5\% and a maximum of 91.5\%. 

After the annotation process, we find that 34\% of the student short answer responses are incorrect (assigned a grade 0). 47\% of student responses are partially correct (grade 0.5), and the remaining 19\% of answers are completely accurate (grade 1). Additionally, the average number of words used in the gold standard answer is 66, while the student responses are more concise, with an average of 36 words per response.

%% file: tables/dataset.tex
\begin{table}[t!]
\centering
\caption{Examples of the SME annotated dataset (best viewed in color).}
\setlength\tabcolsep{4pt}
\setlength\extrarowheight{2pt}
\label{dataset}
\begin{adjustbox}{max width=\linewidth}
\begin{tabular}
{p{3cm}  p{4cm}  p{4cm} P{1cm}}
\toprule
\multicolumn{1}{c}{\textbf{Question}} &  \multicolumn{1}{c}{\textbf{Gold Standard Answer}} & \multicolumn{1}{c}{\textbf{Student Response}} & \multicolumn{1}{c}{\textbf{Grade}} \\
\midrule
Fiber is not really a nutrient, because it passes through our body undigested. Why can't fiber be digested and why is it important to our diet? &
\{\textcolor{blue}{Fiber is a type of carbohydrate that  cannot be digested by the enzymes in the human body}\}. Despite this, it is an important part of our diet because it can help to bulk up our stools, slow down the absorption of sugar and cholesterol, and promote the growth of beneficial bacteria in our gut.

& Fiber promotes \textless \textcolor{ForestGreen}{regular bowel movement by adding bulk}\textgreater, and it \textless \textcolor{ForestGreen}{regulates the blood glucose consumption rate}\textgreater.
It \textless \textcolor{ForestGreen}{removes excess cholesterol from the body}\textgreater.
It binds to the cholesterol in the small intestine, then attaches to the cholesterol and prevents the cholesterol particles from entering the bloodstream
Fiber-rich diets also have \{\textcolor{Dandelion}{a protective role in reducing the occurrence of colon cancer}\}& 0.5 \\ \hline
Fiber is not really a nutrient, because it passes through our body undigested. Why can't fiber be digested and why is it important to our diet? & Fiber is a type of carbohydrate that cannot be digested by the enzymes in human body. Despite this, it is an important part of our diet because \{\textcolor{blue}{it can help to bulk up our stools, slow down the absorption of sugar and cholesterol, and promote the growth of beneficial bacteria in our gut}\}. & Fiber can't be digested because it actually [\textcolor{red}{absorbs water so its important to push waste out of the body}] & 0 \\
\bottomrule

\end{tabular}
\end{adjustbox}
\vspace{2mm}
\end{table}

%% file: sections/benchmark.tex
\subsection{Experimental Setup}

The task of ``Marking", as conceptualized in this study, is approached as a classification problem. Given a student response and a gold standard answer, our objective is to label each word in the student response as either correct, incorrect, or irrelevant. To accomplish this, we employ transformer models, specifically BERT (base/large) \cite{bert} and RoBERTa (base/large) \cite{liu2019roberta}, which have demonstrated remarkable performance in various NLP tasks.

\subsubsection{Classifier Design}

Our classifier design is built upon a pre-trained, masked language model with an inference label classifier head on the top layer. The transformer model, denoted by \(T\), takes a sequence of words \(S = \{w_{0}, w_{1},.., w_{n-1}\}\) as input and outputs embeddings for each token in the sequence. The embeddings are given by \(T(S) = \{\mathbf{e}_{0}, \mathbf{e}_{1},.., \mathbf{e}_{n-1}\}\), where \(\mathbf{e}_i \in \mathbb{R}^D \,, \ (0 \leq i<n-1)\) is the token embedding, and \(D\) is its dimension. 

The inference classifier uses a weight matrix \(\mathbf{W} \in \mathbb{R}^{D \times |\mc{N}|}\) that takes the computed token embeddings as input, where \(\mc{N}\) represents the set of all possible inference classes. The classifier outputs scores for all inference labels for each token in the sentence. Passing these scores through a softmax nonlinearity provides probabilities \(\mathbf{p}_i \in \mathbb{R}^{|\mc{N}|}\) for all inference classes in \(\mc{N}\) for a given token \(i\) in \(S\):

\begin{equation}
    \mathbf{p}_i = \rm{softmax}\Bigl(\mathbf{W}\bigl(\mathbf{e}_{{i}}\bigl)\Bigl).
\end{equation}

The overall model \(\mc{M}\) is then given by:

\begin{equation}
    \mc{M}(T, \mathbf{W}, S, i) = \mathbf{p}_{i}, \, 0 \leq i <2n.
\end{equation}

In our experiments, the sequence of input is the gold answer concatenated with the student response, and the labels are the annotations \(A_i\) for the student response. The label set \(\mc{N}\) consists of five labels: entailment (0), contradiction (1), neutral (2), none of these (3), and a special token (\textless sep\textgreater) (4). The labels are assigned to each word in the student response based on the annotations \(A_i\). The special token (\textless sep\textgreater) is used to separate the gold standard answer and the student response in the input sequence.

\subsubsection{Training Methodology}

The model is trained using the e-SNLI~\cite{camburu2018snli} dataset, which provides premise and hypothesis pairs along with inference labels (entailment, neutral, or contradiction) and word-level annotations for the hypotheses. This dataset is particularly suited for our task as it allows the model to learn the nuances of entailment, contradiction, and neutrality at a word level, which is crucial for ``Marking" task. After training on e-SNLI, we test our model on our marking dataset, with the objective of providing a baseline for future research.

\subsubsection{Data Processing and Label Settings}

In our experimental setup, we treat the gold answer as the premise and the student response as the hypothesis. The task is to predict the inference label of each word in the hypothesis, where the inference label can be either entailment, neutral, contradiction, or none of these. For a given premise sentence of arbitrary $n$ words \(\{p_0, p_1, .. p_n\}\) and a hypothesis sentence of $m$ words \(\{h_0, h_1, .. h_m\}\), we separate them by a special token \texttt{[sep]}. Each word is then assigned a label \(l = \{0,1,2,3,4\}\), where  \texttt{[sep]} has the label 4, words marked as entailed are 0, words marked as contradicted are 1, words marked as neutral are 2, and the rest are marked 3. The task of the transformer is then to predict the labels of each word given the sentence. This setup allows language models to learn the relationship between the student response and the gold standard answer at a granular level, thereby enabling a more detailed and accurate assessment of the student's understanding.

In the context of our classifier design, the sequence \(S\) is the concatenation of the premise (gold standard answer) and the hypothesis (student response), separated by the special token \texttt{[sep]}. Formally, \(S = \{p_0, p_1, ..., p_n, \texttt{[sep]}, h_0, h_1, ..., h_m\}\), where \(p_i\) and \(h_i\) are the words in the premise and hypothesis respectively. 

We consider three different label settings (including one described earlier) to cater to different assessment needs:

\begin{itemize}
    \item \textbf{Generic Setting (0-1-2)}: This is the standard setting where entailment, contradiction, and neutrality are assigned distinct labels (0, 1, and 2 respectively). The rest of the words are assigned the label 3, and the special token \texttt{[sep]} is assigned the label 4. This setting allows for a comprehensive assessment of the student's response.
    
    \item \textbf{Contradiction-focused Setting (0-1-0)}: In this setting, we are primarily interested in identifying the incorrect parts of the student's response. Therefore, both entailment and neutrality are assigned the same label (0), while contradiction is assigned a distinct label (1). The rest of the words are assigned the label 3, and the special token \texttt{[sep]} is assigned the label 4. This setting is useful when the focus of the assessment is to identify misconceptions.
    
    \item \textbf{Error-focused Setting (0-1-1)}: This setting is designed to identify both incorrect and irrelevant parts of the student's response. Therefore, entailment is assigned the label 0, while both contradiction and neutrality are assigned the same label (1). The rest of the words are assigned the label 3, and the special token \texttt{[sep]} is assigned the label 4. This setting is useful when the assessment aims to provide a more stringent evaluation of the student's response.
\end{itemize}

These label settings provide flexibility in the assessment process, allowing educators to choose the setting that best aligns with their assessment objectives.

\subsubsection{Training Parameters}

We use cross-entropy loss to train the classifier. The training parameters are set as follows: learning rate of 2e-5, weight decay of 0.1, warmup ratio of 0.05, Adam optimizer \cite{kingma2014adam}, and 3 epochs. These parameters were chosen based on their performance in previous studies using transformer models for similar NLI tasks.

\subsubsection{Special Preprocessing Contributions}

Our preprocessing step introduces two key contributions to the experimental setup, which are designed to optimize the performance of our model and ensure the consistency of our training and testing procedures.

\noindent\textbf{~Dual Instance Pairing (DIP)~} In the e-SNLI dataset, each premise-hypothesis pair is associated with a single label. However, in the ``Marking" task, words in the student response can exhibit multiple labels simultaneously. This discrepancy between the training and testing conditions could potentially hinder the performance of our model. To address this issue, we introduce a novel preprocessing step where we pair two instances belonging to two different labels as one training sample. We refer to this technique as ``Dual Instance Pairing".

Mathematically, we denote two instances from the e-SNLI dataset as \(P1 \texttt{[sep]} H1\) and \(P2 \texttt{[sep]} H2\), where \(P\) represents the premise, \(H\) represents the hypothesis, and \(\texttt{[sep]}\) is the special token separating the premise and hypothesis. In our preprocessing step, we combine these two instances into a single training sample as \(P1 P2 \texttt{[sep]} H1 H2\). This preprocessing step allows us to leverage the large e-SNLI dataset for transfer learning, while ensuring that our training procedure closely aligns with the conditions of our ``Marking" task.

\input{tables/result1}

\noindent\textbf{~Stop Word Removal~} Our second preprocessing contribution involves the exploration of stopword removal. Stopwords are commonly occurring words in a language (such as ``the", ``a", ``in") that are often filtered out before or after processing text. While they carry little semantic content, their removal can sometimes lead to significant changes in the performance of NLP models.

In the context of our marking task, stop words in the student response could potentially be associated with any of the labels (correct, incorrect, irrelevant). Therefore, their removal could impact the performance of our model. In our experiments, we investigate the effect of stop word removal on the performance of our model, providing valuable insights into the role of stop words in the ``Marking" task.

\input{tables/result2}

\subsection{Experiments and Discussion}
We conduct two main experiments in this study. The first experiment investigates the impact of our two preprocessing steps: Dual Instance Pairing (DIP) and stopword removal. The second experiment establishes baselines for the RoBERTa and Bert models for different label settings. Of the evaluation metrics provided, we emphasize the F1 score over accuracy due to varying datapoint numbers in each class as label settings changes. Accuracy is only included for comprehensiveness.

\subsubsection{Impact of Preprocessing Steps}

In the first experiment, we train the RoBERTa-large model on the e-SNLI dataset and test it on BioMarking. 
We evaluate the performance under all three label settings: Generic, Contradiction-focused, and Error-focused. Our results in table~\ref{tab:result1} indicate that both DIP and stopword removal preprocessing steps contribute positively to the model performance.

For instance, in the Generic setting, the RoBERTa-large model with both DIP and stopword removal achieves an F1 score of 0.423. When we remove the DIP preprocessing step, the F1 score drops to 0.394, indicating a decrease in performance. Similarly, when we remove the stopword removal preprocessing step, the F1 score decreases to 0.332.

These results suggest that DIP and stopword removal are beneficial preprocessing steps that enhance the model's ability to accurately assess student responses.

\subsubsection{Performance of Different Models and Label Settings}

In the second experiment, we train RoBERTa (base and large) and BERT (base and large) models on the e-SNLI dataset and test them on BioMarking dataset, as shown in Table~\ref{tab:result2}. We evaluate the performance of these models under all three label settings: Generic, Contradiction-focused, and Error-focused. 

\textbf{Finding 1:} Our results indicate that the RoBERTa-large model outperforms the other models in all three settings. For instance, in the Generic setting, the RoBERTa-large model achieves an F1 score of 0.423, while the RoBERTa-base, BERT-large, and BERT-base models achieve F1 scores of 0.376, 0.379, and 0.331, respectively. This finding is consistent with the literature, which generally reports superior performance of RoBERTa models over BERT models in NLI tasks. 

\textbf{Finding 2:} We find that the Error-focused setting yields the best performance across all models. For the RoBERTa-large model, the F1 scores for the Generic, Contradiction-focused, and Error-focused settings are 0.423, 0.563, and 0.762, respectively. These results suggest that the Error-focused setting, which aims to identify both incorrect and irrelevant parts of the student's response, provides a more stringent and comprehensive evaluation of the student's understanding. 

The superior performance of the Error-focused setting can be attributed to its unique approach of treating both incorrect and irrelevant parts of the student's response as errors. This approach aligns well with educational objectives, as it encourages students to provide precise and relevant responses. 

Firstly, by marking incorrect parts as errors, the model helps identify and rectify misconceptions in students' understanding. This behavior is crucial in education since unaddressed misconceptions can hinder the learning process.
Secondly, by marking irrelevant parts as errors, the model discourages students from including extraneous information in their responses. This behavior is important as it promotes clarity and precision in student responses, which are critical skills in effective communication.

Therefore, the Error-focused setting not only provides a more stringent evaluation of the student's understanding but also promotes critical educational objectives. This finding underscores the potential of our proposed label settings in enhancing the effectiveness of automated grading systems in education.

%% file: tables/result1.tex
\begin{table}[t!]
\centering
\setlength\tabcolsep{4pt}
\setlength\extrarowheight{2pt}
\caption{Performance of the RoBERTa-large model on the ``Marking" task under different preprocessing conditions and label settings. The results demonstrate the positive impact of the Dual Instance Pairing (DIP) and stopword removal preprocessing steps on the model performance.}
\label{tab:result1}
\begin{adjustbox}{max width=\linewidth}
\begin{tabular}{ccccccccc}
\toprule
\multicolumn{3}{c}{\textbf{Label}} & \multirow{2}{*}{\textbf{Stopwords}} & \multirow{2}{*}{\textbf{Pairs}} & \multirow{2}{*}{\textbf{Precision} $\uparrow$} & \multirow{2}{*}{\textbf{Recall} $\uparrow$} & \multirow{2}{*}{\textbf{F1} $\uparrow$} & \multirow{2}{*}{\textbf{Accuracy} $\uparrow$} \\
\textbf{Ent} & \textbf{Con} & \textbf{Neu} &  &  &  &  &  &  \\
\midrule
\multirow{4}{*}{\textbf{0}} & \multirow{4}{*}{\textbf{1}} & \multirow{4}{*}{\textbf{2}} & \textbf{Yes} & \textbf{Yes} & \underline{0.425} &  \underline{0.421} & \underline{0.423} & 0.625 \\			
 &  &  & \textbf{Yes} & \textbf{No} &  0.407 & 0.382 & 0.394 & \underline{0.638} \\	
 &  &  & \textbf{No} & \textbf{Yes} & 0.303 & 0.368 & 0.332 & 0.609 \\					
 &  &  & \textbf{No} & \textbf{No} & 0.332 & 0.327 & 0.33 & 0.613 \\ 
 \midrule
\multirow{4}{*}{\textbf{0}} & \multirow{4}{*}{\textbf{1}} & \multirow{4}{*}{\textbf{0}} & \textbf{Yes} & \textbf{Yes} & 0.819 & \underline{0.429} & \underline{0.563} & 0.904 \\ 			
 &  &  & \textbf{Yes} & \textbf{No} & \underline{0.918} & 0.399 & 0.556 & \underline{0.908}  \\
 &  &  & \textbf{No} & \textbf{Yes} & 0.557 & 0.419 & 0.478 & 0.885 \\ 			
 &  &  & \textbf{No} & \textbf{No} & 0.701 & 0.303 & 0.423 & 0.896 \\		
 \midrule
\multirow{4}{*}{\textbf{0}} & \multirow{4}{*}{\textbf{1}} & \multirow{4}{*}{\textbf{1}} & \textbf{Yes} & \textbf{Yes} & \underline{0.747} & 0.778 & \underline{0.762} & 0.704 \\ 	

 &  &  & \textbf{Yes} & \textbf{No} & 0.718 & \underline{0.805} & 0.756 & \underline{0.708} \\  		
 &  &  & \textbf{No} & \textbf{Yes} & 0.702  & 0.796 & 0.746 & 0.690 \\
 &  &  & \textbf{No} & \textbf{No} &  0.709 & 0.76 & 0.734 & 0.685 \\
 \bottomrule
\end{tabular}
\end{adjustbox}
\end{table}

%% file: tables/result2.tex
\begin{table}[t!]
\centering
\setlength\extrarowheight{2pt}
\setlength\tabcolsep{4pt}
\caption{Performance of different models (RoBERTa and BERT, base and large) on the ``Marking" task under three label settings: Generic (0-1-2), Contradiction-focused (0-1-0), and Error-focused (0-1-1). The table reports the evaluation scores for each combination of model and label setting. The results highlight the superior performance of the RoBERTa-large model and the effectiveness of the Error-focused label setting.}
\label{tab:result2}
\begin{adjustbox}{max width=\linewidth}
\begin{tabular}{cccccccc}
\toprule
\multirow{2}{*}{\textbf{Model}} & \multicolumn{3}{c}{\textbf{Label}} & \multirow{2}{*}{\textbf{Precision} $\uparrow$} & \multirow{2}{*}{\textbf{Recall} $\uparrow$} & \multirow{2}{*}{\textbf{F1} $\uparrow$} & \multirow{2}{*}{\textbf{Accuracy} $\uparrow$} \\ 
 & \textbf{Ent} & \textbf{Con} & \textbf{Neu} &  &  &  &  \\
 \midrule
\multirow{3}{*}{\textbf{Roberta-large}} & \textbf{0} & \textbf{1} & \textbf{2} & 0.425 & 0.421 & 0.423 & 0.625 \\
 & \textbf{0} & \textbf{1} & \textbf{0} & 0.819 & 0.429 & 0.563 & 0.904 \\		
 & \textbf{0} & \textbf{1} & \textbf{1} & 0.747 & 0.778 & 0.762 & 0.704 \\ 			
\midrule
\multirow{3}{*}{\textbf{Roberta-base}} & \textbf{0} & \textbf{1} & \textbf{2} & 0.380 & 0.372 & 0.376 & 0.611 \\ 						
 & \textbf{0} & \textbf{1} & \textbf{0} & 0.684 & 0.43 & 0.528 & 0.889 \\ 			
 & \textbf{0} & \textbf{1} & \textbf{1} & 0.721 & 0.774 & 0.746 & 0.699 \\				
 \midrule
\multirow{3}{*}{\textbf{Bert-large}} & \textbf{0} & \textbf{1} & \textbf{2} & 0.405 & 0.402 & 0.379 & 0.593 \\ 			
 & \textbf{0} & \textbf{1} & \textbf{0} & 0.577 & 0.462 & 0.513 & 0.873 \\
 & \textbf{0} & \textbf{1} & \textbf{1} & 0.718 & 0.741 & 0.729 & 0.685 \\		
 \midrule
\multirow{3}{*}{\textbf{Bert-base}} & \textbf{0} & \textbf{1} & \textbf{2} & 0.292 & 0.384 & 0.331 & 0.564 \\ 						
 & \textbf{0} & \textbf{1} & \textbf{0} & 0.531 & 0.423 & 0.471 & 0.863 \\			
 & \textbf{0} & \textbf{1} & \textbf{1} & 0.728 & 0.695 & 0.711 & 0.679 \\					
 \bottomrule
\end{tabular}
\end{adjustbox}
\end{table}

%% file: sections/conclusion.tex
In this paper, we introduced the task of ``Marking" as a novel approach to automated grading in educational settings.
By leveraging AI, we aimed to move beyond binary assessments and provide a more nuanced and detailed form of feedback, similar to what is provided by experienced educators. 
Our approach involved evaluating student responses against a gold standard, meticulously highlighting text segments as correct, incorrect, or irrelevant, and identifying key concepts or information the student omitted. 
Our primary contribution is the introduction of a novel dataset, BioMarking, meticulously curated by Subject Matter Experts for the ``Marking" task. 
This dataset also serves as a benchmark for the AI in the education community to engage with and improve upon. 
We also presented initial baseline performance using state-of-the-art NLP techniques, adapted to ``Marking" task. 
While these initial results are promising, they also highlight the complexity and challenge of the ``Marking", setting a clear path for future research.
Additionally, the task of ``Marking" holds significant potential for enhancing the learning experience for students. 
By providing detailed feedback on their responses, it can help students identify their strengths and weaknesses, gain a deeper understanding of the subject matter, and guide their learning process. 
For educators, it offers a scalable solution for grading, reducing their workload while ensuring a high level of feedback quality. 
We believe that our dataset will stimulate further research in this area, leading to the development of more sophisticated models capable of providing detailed and nuanced feedback to students.